\documentclass[conference]{IEEEtran}
\IEEEoverridecommandlockouts
\usepackage{cite}
\usepackage[colorlinks,
            linkcolor=green,
            anchorcolor=green,
            citecolor=green]
            {hyperref}
\usepackage{amsmath,amssymb,amsfonts}
\usepackage{algorithmic}
\usepackage{graphicx}
\usepackage{textcomp}
\usepackage{xcolor}
\usepackage{subfig}
\usepackage{booktabs}
\usepackage{comment}
\usepackage{caption}
\usepackage{multicol}
\def\BibTeX{{\rm B\kern-.05em{\sc i\kern-.025em b}\kern-.08em
    T\kern-.1667em\lower.7ex\hbox{E}\kern-.125emX}}
\begin{document}

\title{Adapting Image-to-Video Diffusion Models for Large-Motion Frame Interpolation}

\author{\IEEEauthorblockN{\textsuperscript{} Luoxu Jin}
\IEEEauthorblockA{\textit{\textit{CSCE, Graduate School of FSE}} \\
\textit{Waseda University} \\
Tokyo, Japan \\
}
\and
\IEEEauthorblockN{\textsuperscript{} Hiroshi Watanabe}
\IEEEauthorblockA{\textit{\textit{CSCE, Graduate School of FSE}} \\
\textit{Waseda University} \\
Tokyo, Japan \\
}
}
\maketitle

\begin{abstract}
With the development of video generation models has advanced significantly in recent years, we adopt large-scale image-to-video diffusion models for video frame interpolation. We present a conditional encoder designed to adapt an image-to-video model for large-motion frame interpolation. To enhance performance, we integrate a dual-branch feature extractor and propose a cross-frame attention mechanism that effectively captures both spatial and temporal information, enabling accurate interpolations of intermediate frames. Our approach demonstrates superior performance on the Fréchet Video Distance (FVD) metric when evaluated against other state-of-the-art approaches, particularly in handling large motion scenarios, highlighting advancements in generative-based methodologies.
\end{abstract}


\begin{figure*}[!t] 
  \centering
  \includegraphics[width=0.9\textwidth]{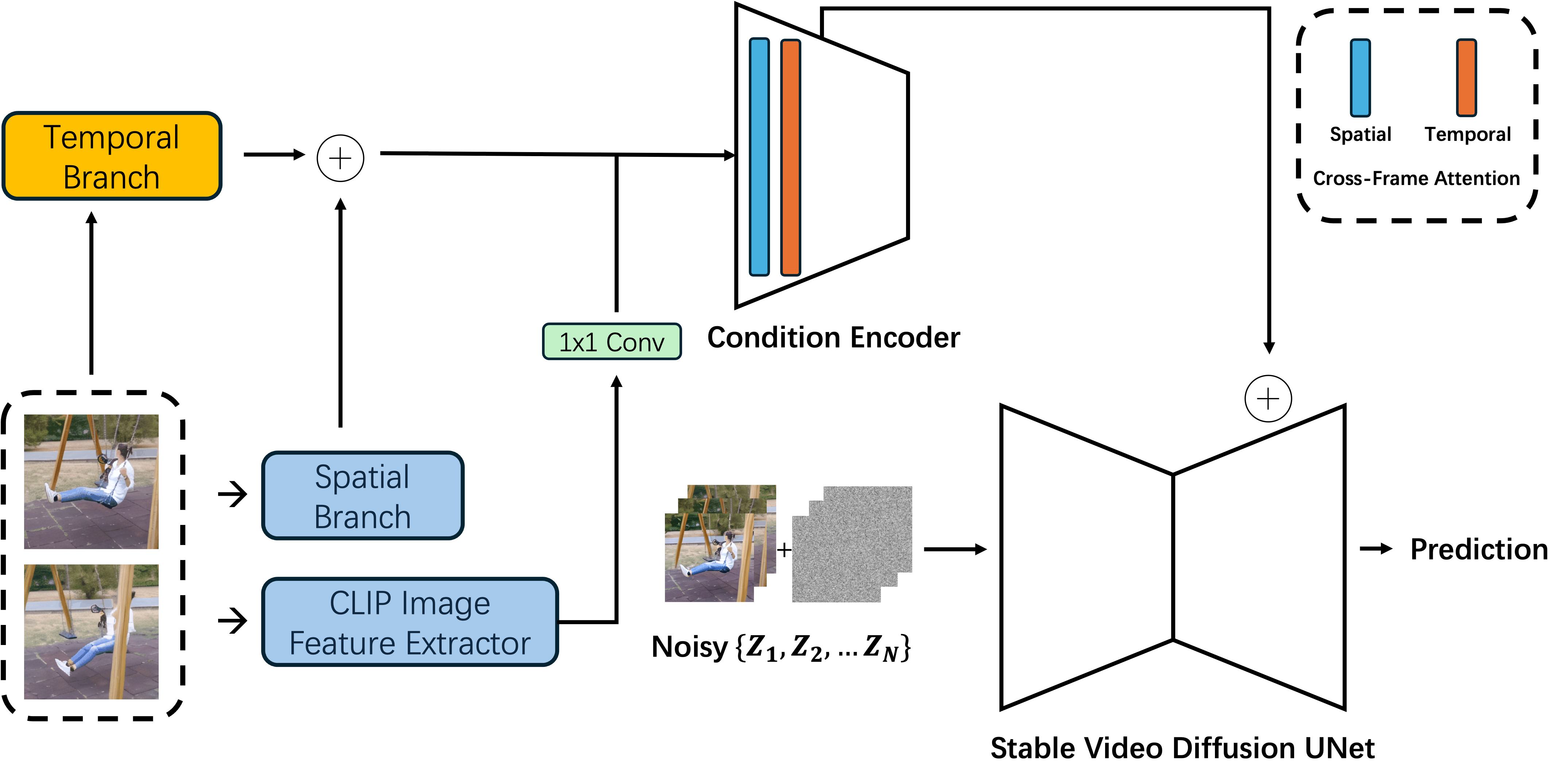} 
  \caption{The input video data \(\{I_1, I_2, \dots, I_N\}\) is first encoded into latent representations, where Gaussian noise is added to produce the noisy representations \(\{z_1, z_2, \dots, z_N\}\). Subsequently, the spatial and temporal features of the first and last frames are extracted through separate branches and fed into the conditional encoder. The conditional encoder integrates these features while employing zero initialization to ensure stable training. Finally, the integrated features are incorporated into a pre-trained 3D U-Net.
} 
  \label{fig:framwork} 
\end{figure*}

\section{Introduction}
Video frame interpolation (VFI) refers to the process of generating intermediate frames between two existing frames in a video to increase the frame rate or to smooth out motion. This technique is widely used in a variety of applications, such as creating slow-motion videos, enhancing video quality for smoother playback, or improving the visual experience in tasks like animation. Video frame interpolation has traditionally been addressed using convolutional neural networks and flow-based models, which synthesize intermediate frames using the convolution kernel to capture local and long-range dependencies in video data. These methods typically rely on explicit motion estimation techniques, such as optical flow, to predict the motion field between consecutive frames, which is then used to guide the interpolation process\cite{Cheng2020VideoFI},\cite{Danier2021STMFNetAS},\cite{Jiang2017SuperSH},\cite{Niklaus2020SoftmaxSF},\cite{Sim2021XVFIEV},\cite{Kong2022IFRNetIF}. However, they often exhibit limitations in handling large, non-linear motions, occlusions, or complex object deformations due to the inherent constraints of motion estimation algorithms and the locality of convolution kernel. 

In contrast, generative models, such as diffusion-based approaches, represent a significant advancement in learning implicit motion representations and scene dynamics directly from large-scale video datasets. These models utilize rich prior knowledge to synthesize temporally and spatially coherent frames, even in cases involving substantial motion or ambiguous scenarios. By modeling the frame interpolation task probabilistically, generative approaches can produce multiple plausible intermediate frames, offering greater flexibility and robustness compared to traditional deterministic methods\cite{LDMVFI},\cite{VIDM},\cite{SVDKFI},\cite{framer}. 

Training a video diffusion model for video frame interpolation is a resource-intensive process that demands significant computational power to capture and model complex temporal dynamics between frames. To address this issue, we propose leveraging pre-trained image-to-video diffusion models and adapting them specifically for video frame interpolation tasks. Our approach is similar to Parameter-Efficient Fine-Tuning (PEFT) techniques, enabling effective optimization of the base model while minimizing the need for extensive retraining.

Building on the success of ControlNet\cite{controlnet} in the text-to-image domain, which has demonstrated exceptional control capabilities, we extend this method to keyframe-guided video frame interpolation. Experimental results indicate that this adaptation of ControlNet achieves great performance, maintaining temporal and spatial consistency while addressing the complexities of interpolation tasks. Moreover, to enhance temporal consistency in the interpolated frames, we integrate bidirectional optical flow and depth map into our approach, ensuring smoother and more coherent transitions between frames. Additionally, we incorporate a cross-frame attention mechanism within each transformer block, allowing the model to better capture spatial and temporal representations by emphasizing cross-frame feature. These enhancements improve the quality of the interpolated frames, enabling the model to handle complex motion dynamics and achieve better performance in video frame interpolation tasks. 

We evaluated our interpolation model across a diverse range of cases, including real-world styles, linear sketches, and anime styles, demonstrating its impressive performance. Furthermore, our method is evaluated on datasets featuring scenarios where the first and last frames exhibit large motion, addressing ambiguous motion cases compared to previous deterministic-based methods. These results indicate that incorporating the plug-and-play conditional encoder enhances the model's versatility and adaptability for video frame interpolation tasks. In summary, our contributions are as follows. 

\begin{enumerate}
    \item We introduce a conditional encoder that adapts an image-to-video model to handle large-motion scenarios in video frame interpolation.
    \item We propose a dual-branch architecture to capture spatial-temporal features and integrate a cross-frame attention module into each transformer block, enhancing interpolation performance.
\end{enumerate}

\section{Related Work}
\subsection{Video Frame Interpolation}
The frame interpolation task is defined as the process of synthesizing intermediate frames given two input frames, enabling smoother transitions and higher temporal resolution in videos. This task has been widely explored, with a significant body of research focusing on flow-based methods, which estimate optical flow between the input frames to guide the generation of intermediate frames\cite{Jiang2017SuperSH}\cite{Niklaus2020SoftmaxSF}. 

Alongside these, other approaches employ CNNs or transformers\cite{Attention} to learn feature representations directly from the input frames, bypassing explicit motion estimation. These models extract spatial and temporal features and synthesize the intermediate frame in a data-driven manner, demonstrating versatility and efficiency\cite{AMT},\cite{FILM},\cite{RIFE}. However, previous work often assumes that the two input frames are closely spaced, which limits their ability to interpolate ambiguous or complex motion accurately. This assumption makes traditional methods less effective for scenarios with large temporal gaps or large motion dynamics. 

Recently, generative models have demonstrated remarkable potential in video synthesis tasks and have shown significant promise for applications in video frame interpolation \cite{VIDM}, \cite{SVDKFI}, \cite{LDMVFI}, \cite{framer}. In this context, we explore the use of video diffusion models for frame interpolation by incorporating a conditional encoder. This approach facilitates the adaptation of pre-trained generative models for interpolation tasks, reducing computational overhead while preserving high-quality synthesis.

\subsection{Diffusion Model for Image and Video Generation}
Diffusion models are a type of generative model that have demonstrated remarkable capabilities in the text-to-image domain, achieving state-of-the-art performance among generative approaches \cite{ddpm},\cite{ddim},\cite{diffusion_beat_gan},\cite{cfg},\cite{ldm}. Building on their success in image generation, recent research has increasingly focused on applying diffusion models to video generation\cite{SVD},\cite{CogVideo}. Some approaches involve training a plug-and-play motion module that integrates seamlessly with existing text-to-image diffusion models\cite{animediff}, while others extend pre-trained text-to-image generation models by adding temporal modules to handle video dynamics\cite{Align},\cite{make-a-video}. Both methods have shown promising results in synthesizing temporally coherent and visually compelling video content.

\section{Proposed Method}
\subsection{Preliminaries}
Diffusion models operate through two fundamental phases that support their generative capabilities. The forward diffusion process involves progressively corrupting real data, such as an image, by adding noise over a series of steps until the data is transformed into pure noise. In contrast, the reverse diffusion process trains the model to iteratively denoising this noisy data step by step, reconstructing coherent and high-quality samples from random noise. 

In this research, we utilize Stable Video Diffusion (SVD)\cite{SVD} as our base model, a latent diffusion framework akin to Stable Diffusion\cite{ldm}, tailored for video generation tasks. Given a sequence of $N$ video frames \{$I_1,I_2,...,I_N$\} the pre-trained autoencoder $\mathcal{E}$ encodes the frames into a latent representation \{$z_1,z_2,..,z_N$\} During training, a standard diffusion process is applied in the latent space, where gaussian noise $\epsilon \sim N(0,I)$ is added to the latent representations $z_t$. Here, $z_t$ is represented as $z_t = \alpha_t z + \sigma_t \epsilon$ at time step t, where $\alpha_t$ and $\sigma_t$ are predefined by the noise schedule. In stable video diffusion, v-prediction is employed, where $v = \alpha_t z - \sigma_t \epsilon$. Here, $\tau_\theta$ is a conditional encoder that encodes the condition $c$ into the model $f_\theta$. The model $f_\theta$ is trained using the following loss function:
\begin{equation}
    \mathcal{L} = \mathbb{E}_{\epsilon \sim N(0,I), t \sim Uniform(1,T)} \, ||f_\theta (z_t, t, \tau_\theta(c)) - v||.
\end{equation}
At inference time, the reverse diffusion process begins with pure gaussian noise $z \sim N(0,I)$ and iteratively denoising it to generate a sequence of latent representations. These latent representations are then passed through the decoder $\mathcal{D}$, which reconstructs the corresponding frames in the image space. 

\begin{figure}[!t]
    \centering
    \includegraphics[width=\linewidth]{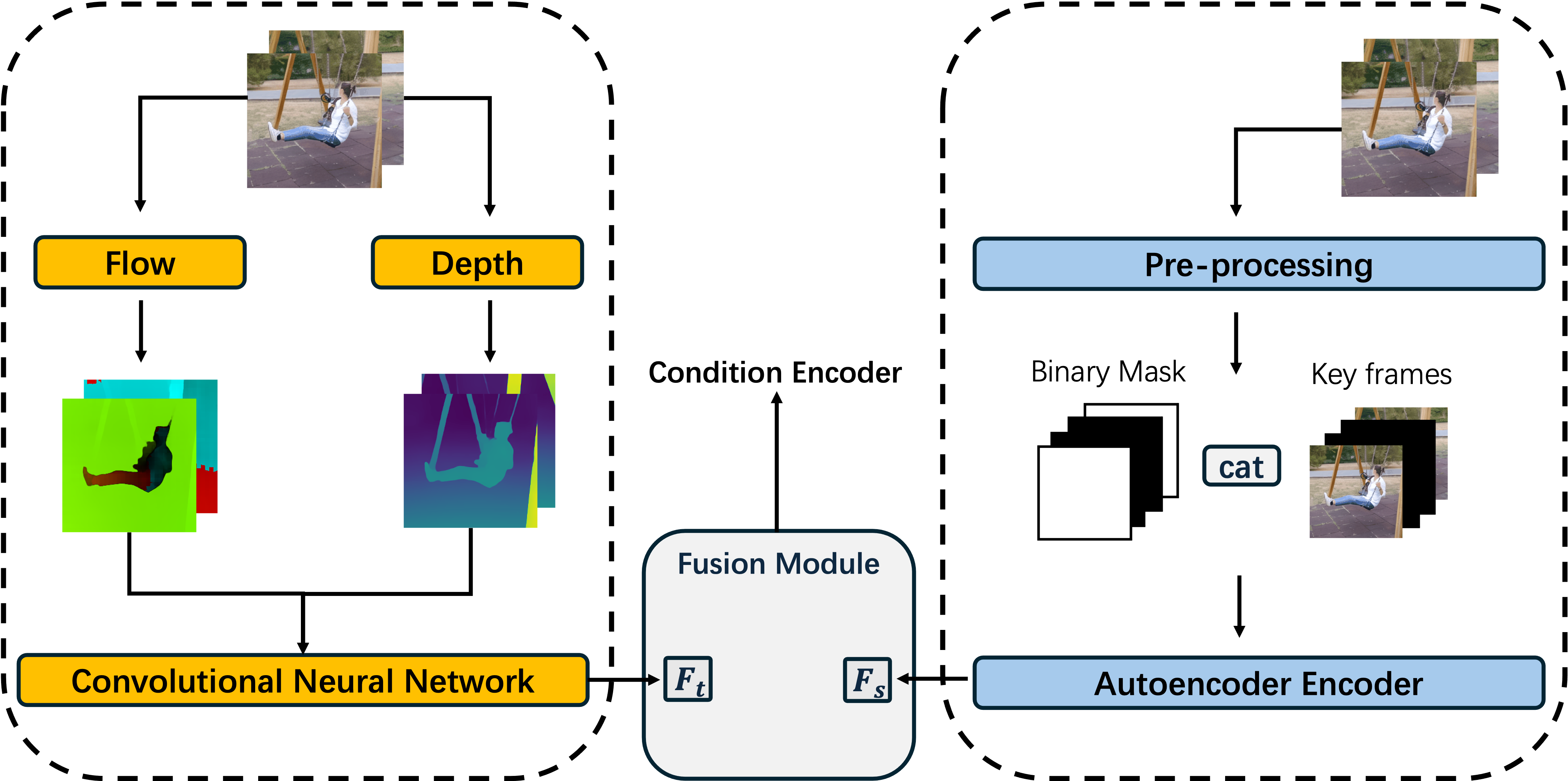}
    \caption{The first and last frames are processed through distinct feature extractors, followed by a fusion process to integrate their features.}
    \label{fig:fusion}
\end{figure}

\begin{figure}[!t]
    \centering
    \includegraphics[width=0.9\linewidth]{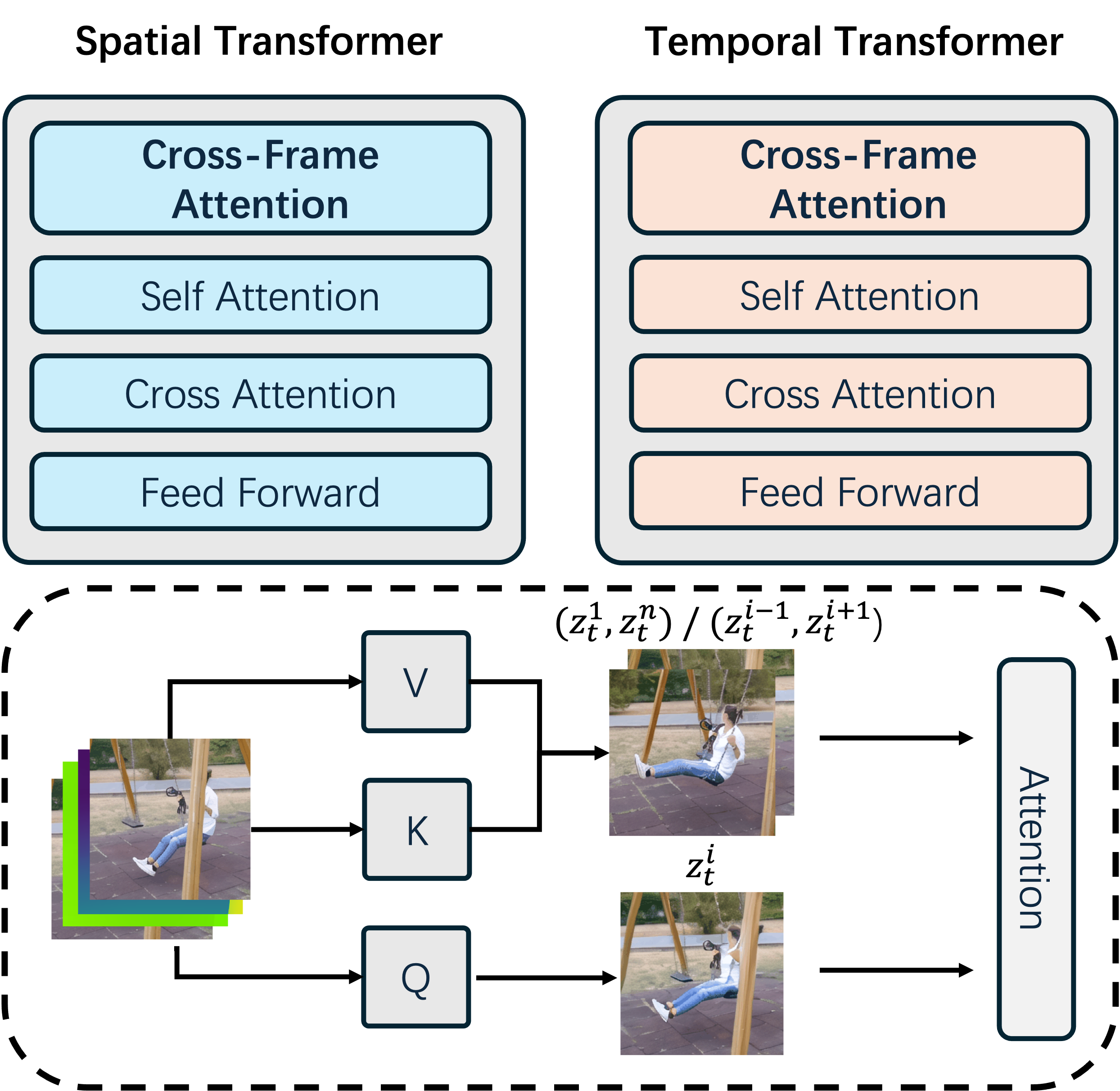} 
    \caption{In both the spatial transformer block and the temporal transformer block, a cross-frame attention block is first integrated to enhance feature representation.} 
    \label{fig:attn}
\end{figure}

\begin{figure*}[!t]
    \centering
    \includegraphics[width=\textwidth]{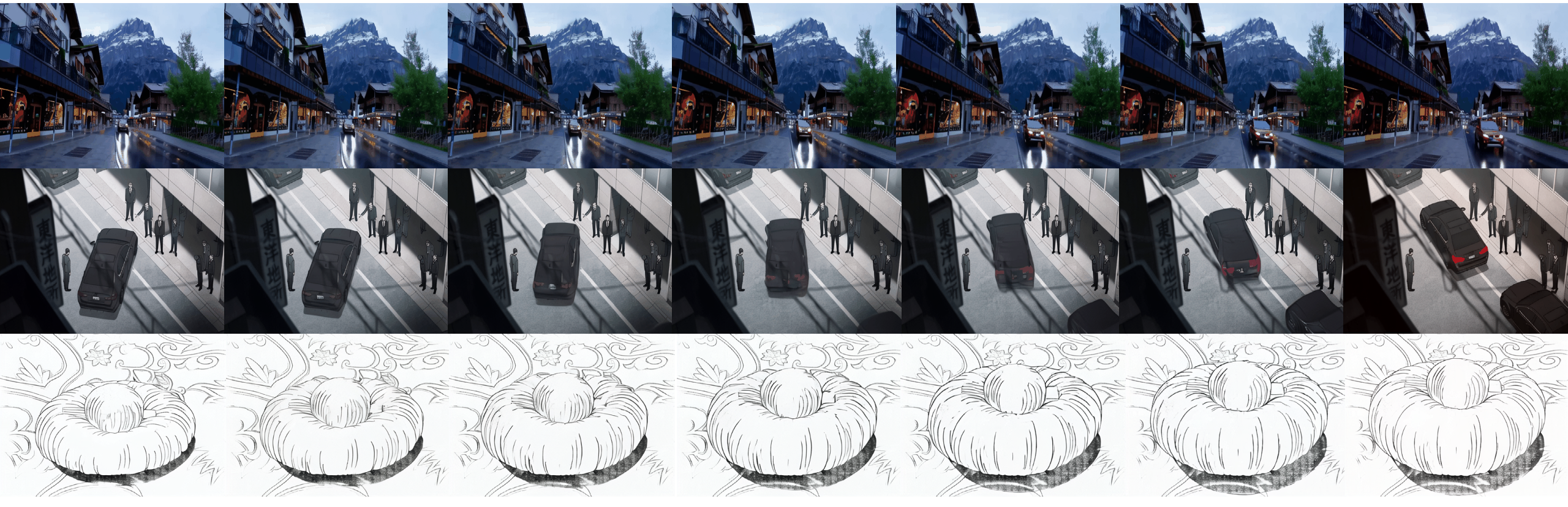} 
    \caption{Interpolation results across different styles. From the first row to the last row, the results correspond to real-world, anime, and sketch styles.}
    \label{fig:style}
\end{figure*}

\begin{table*}[!t]
\centering
\caption{Comparison of results on the DAVIS-7 and UCF101-7 datasets, evaluated on all 7 generated frames.}
\setlength{\tabcolsep}{7pt} 
\begin{tabular}{lcccccc|lcccccc}
\toprule
\multicolumn{6}{c}{\textbf{DAVIS-7 Dataset}} & & \multicolumn{6}{c}{\textbf{UCF101-7 Dataset}} \\
\cmidrule{1-7} \cmidrule{8-13}
& PSNR $\uparrow$ & SSIM $\uparrow$ & LPIPS $\downarrow$ & FID $\downarrow$ & FVD $\downarrow$ & &
& PSNR $\uparrow$ & SSIM $\uparrow$ & LPIPS $\downarrow$ & FID $\downarrow$ & FVD $\downarrow$ \\
\midrule
AMT\cite{AMT}        & \textbf{21.38} & \textbf{0.5881} & 0.2540 & 22.16 & 279.1 & &
AMT\cite{AMT}        & \textbf{26.44} & \textbf{0.8428} & 0.1712 & 19.66  & 296.0 \\
FILM\cite{FILM}      & 20.63          & 0.5489          & \textbf{0.2421} & \textbf{10.89} & 282.5 & &
FILM\cite{FILM}      & 25.99          & 0.8309          & \textbf{0.1546} & \textbf{10.27} & 321.2 \\
SVDKFI\cite{SVDKFI}  & 15.47          & 0.3226          & 0.4035 & 30.55 & 503.0 & &
SVDKFI\cite{SVDKFI}  & 18.91          & 0.6520          & 0.2897 & 56.41 & 361.4 \\
Ours                 & 19.53          & 0.4915          & 0.3064 & 24.84 & \textbf{177.79} & &
Ours                 & 23.09          & 0.7541          & 0.2501 & 38.68  & \textbf{265.74} \\
\bottomrule
\end{tabular}
\label{tab:combined}
\end{table*}

\subsection{Conditional Encoder}
We present a novel plug-and-play conditional encoder module designed specifically for video frame interpolation, As discussed in previous work\cite{sparse}, adding noised latents to the conditional encoder can cause the model to overlook these samples, leading to inconsistencies in frame interpolation during training. Therefore, we also eliminate the noised latents and instead rely solely on the conditional inputs as shown in Fig \ref{fig:framwork}. Firstly, we extract the first and last frames from the video data \(\{I_1, I_2, \dots, I_N\}\), which results in \(\{I_1, I_N\}\). We design spatial and temporal branches to extract spatial-temporal features, aiming to enhance interpolation performance shown in Fig. \ref{fig:fusion}. Additionally, a \(1 \times 1\) convolutional layer is employed to integrate conditional image CLIP \cite{clip} features for further processing.

\textbf{Spatial Branch:} Zero images are inserted between the first and last frames to represent the unconditional phase, with a concatenated binary mask to distinguish phases. The binary mask is set as \(m = 1\) for the conditional phase in the first and last frames, and \(m = 0\) for the unconditional phase in intermediate frames. These components are concatenated along the channel dimension. A pre-trained autoencoder is then utilized to encode the combined inputs, extracting spatial features. 

\textbf{Temporal Branch:} A pre-trained optical flow extractor \cite{RAFT} and a depth extractor \cite{Depth} are utilized to compute bi-directional optical flow and the keyframe depth map. These motion representations are subsequently processed through a convolutional network to produce temporal features. The resulting temporal features indicate keyframe occlusions, aiding in the interpolation modeling process.

We adopt a simple early fusion strategy to combine spatial and temporal features, as the fused features will be processed by a complex conditional encoder capable of effectively learning the required representations autonomously. Specifically, the fusion of features is defined by the following equation:
\begin{equation}
\mathbf{F_s} = \text{CBAM}(\mathbf{F_s}), \quad \mathbf{F_t} = \text{CBAM}(\mathbf{F_t}),
\end{equation}
\begin{equation}
\mathbf{F_{used}} = \text{CrossAttention}(\mathbf{F_s} \cdot \mathbf{W}_Q, \mathbf{F_t} \cdot \mathbf{W}_K, \mathbf{F_t} \cdot \mathbf{W}_V)
\end{equation}
\begin{equation}
\mathbf{F}_\text{used} = \mathbf{F}_\text{used} + \mathbf{F}_\text{s}
\end{equation}
Where \(F_s\) and \(F_t\) denote the spatial and temporal features, respectively. The spatial and temporal features \(F_s\) and \(F_t\) are passed through the CBAM\cite{CBAM} to obtain corresponding weights, which are then applied adaptively to the spatial-temporal features. These combined features are fed into the Cross Attention for feature fusion. This approach adaptively balances the contributions of spatial and temporal features in the fused representation.

The design of the conditional encoder is inspired by ControlNet\cite{controlnet}. It replicates the encoder part of the pre-trained Stable Video Diffusion U-Net, initializing the weights identically. The conditional encoder comprises a convolutional block, a spatial transformer, and a temporal transformer block. In the spatial and temporal transformer blocks, we introduce cross-frame attention to enhance the representation of both spatial and temporal fused features. The attention layer processes \(n\) inputs, denoted as \(z = [z^1, \ldots, z^n]\), where \(i = 1, \ldots, n\). Linear projection operations are applied to generate query, key, and value features. In the spatial transformer block, for each query feature derived from a frame \(z_i\), we utilize the first and last frames as keyframes, denoted as \(z_{1}\) and \(z_{n}\). These features are concatenated to form \(K^{(1, n)}\) and \(V^{(1, n)}\), respectively. The cross-frame attention mechanism, \(CrossFrameAttn(Q, K, V)\), is mathematically defined as follows.
\begin{equation}
   \text{Softmax}\left(\frac{Q \cdot ({K^{(1,n)}})^\top}{\sqrt{d_k}}\right) V^{(1,n)}
\end{equation}
Similarly, in the temporal transformer block, for each query feature derived from a frame \(z_i\), as well as the preceding and succeeding frames, \(z_{i-1}\) and \(z_{i+1}\), these features are concatenated to form \(K^{(i-1, i+1)}\) and \(V^{(i-1, i+1)}\), respectively. The cross-frame attention mechanism, \(CrossFrameAttn(Q, K, V)\), is mathematically defined as follows.
\begin{equation}
   \text{Softmax}\left(\frac{Q \cdot ({K^{(i-1,i+1)}})^\top}{\sqrt{d_k}}\right) V^{(i-1,i+1)}
\end{equation}

Specifically, in the spatial transformer block, conditional frames are utilized as key-value pairs to propagate keyframe appearance correlations from the spatial features. In the temporal transformer block, cross-frame interactions are modeled using conditional frames as key-value pairs to propagate implicit temporal coherence across other features, as illustrated in Fig. \ref{fig:attn}.

\section{Experiment}
\subsection{Experiment Settings}
We implement our conditional encoder upon Stable Video Diffusion model, an image-to-video model trained on the WebVid10M\cite{webvid10m} large-scale video dataset. For training, we utilize the Vimeo-90k Septuplet\cite{vimeo90k} dataset as the training set. Due to computational resource constraints, we employ the 8-bit AdamW optimizer with a learning rate of \(1 \times 10^{-5}\). The performance of our model is evaluated on the DAVIS-7\cite{DAVIS} and UCF101-7\cite{UCF101} datasets using a range of metrics, including PSNR, SSIM, LPIPS\cite{lpips}, FID\cite{fid}, and FVD\cite{fvd}.

\begin{figure*}[!t]
    \centering
    \includegraphics[width=\textwidth]{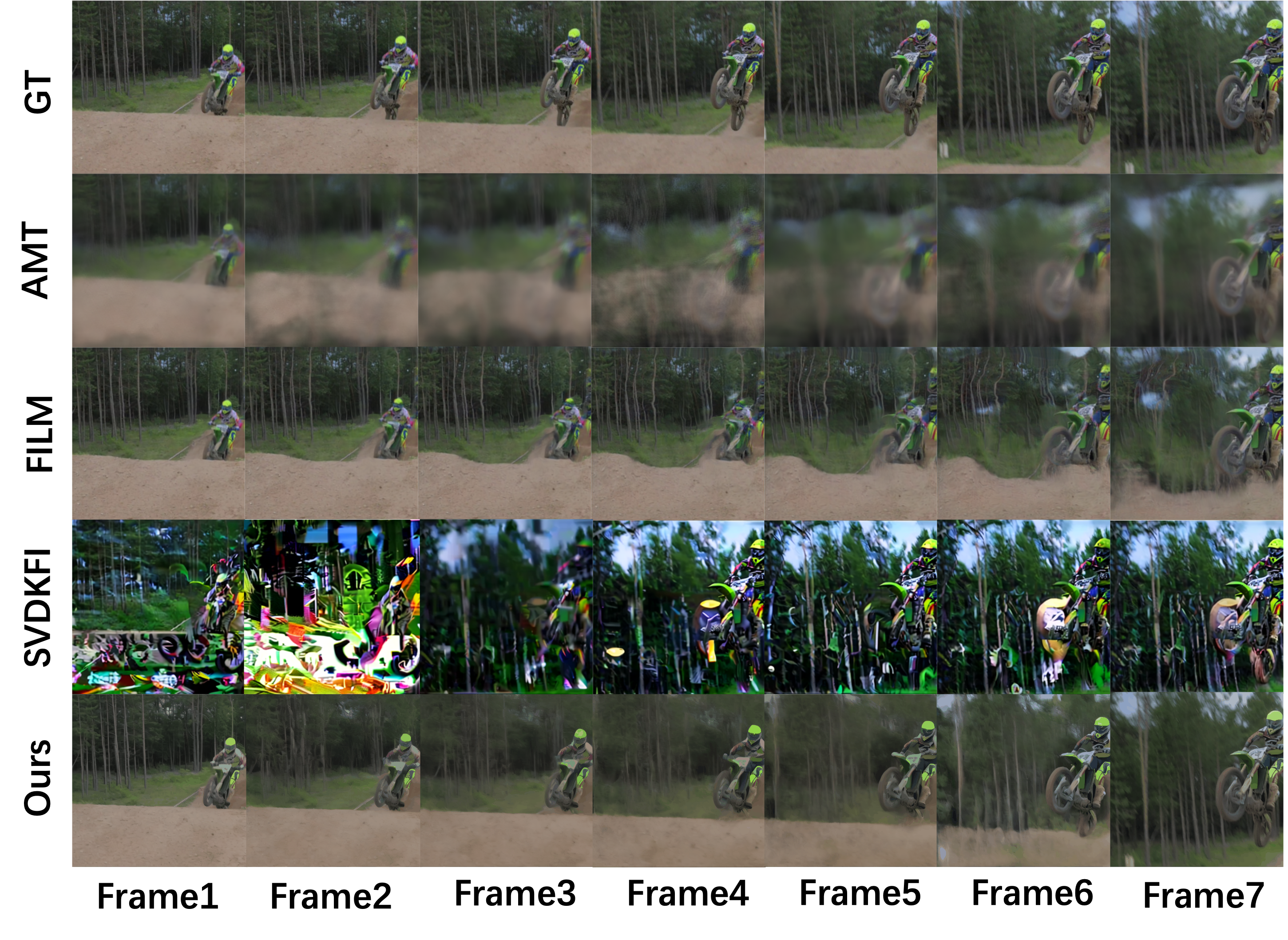} 
    \caption{The example demonstrates a large-motion scenario where traditional methods fail to predict the direction and magnitude of motion accurately, often resulting in blurred or inconsistent intermediate frames. In contrast, our approach reconstruct realistic motion trajectories, producing coherent intermediate frames that capture accurate motion dynamics.}
    \label{fig:quality}
\end{figure*}

\subsection{Quantitative Result}
We select several comparison methods, including AMT\cite{AMT}, FILM\cite{FILM}, and SVDKFI\cite{SVDKFI}. AMT and FILM represent traditional video frame interpolation methods, while SVDKFI and our method are generative-based approaches. The best results are highlighted in bold font. As discussed in VIDIM\cite{VIDM}, metrics such as PSNR, SSIM, LPIPS, and FID are not fully suitable for evaluating video frame interpolation as they may not align with human perceptual judgment. Our method achieves the best performance on the FVD metric, as shown in Table \ref{tab:combined}.

\begin{table}[!t]
    \centering
    \caption{Ablation study of proposed components, evaluating the middle of 7 generated frames at 512 × 512 resolution}
    \setlength{\tabcolsep}{5pt} 
    \begin{tabular} {@{}lcccccc@{}}
        \toprule
        & PSNR↑ & SSIM↑ & LPIPS↓ & FID↓ & FVD↓ \\
        \midrule
        Baseline & 19.01  &  0.4839  & 0.3502 & 24.52 & 183.1 \\
        w/ Cross-frame attention & 18.86 & 0.4794 & 0.3541 & 24.35 & 157.3  \\
        w/ Temporal features & 19.03 & 0.4869 & 0.3520 & 24.69 & 169.4 \\
        Ours & \textbf{19.18} & \textbf{0.4874} & \textbf{0.3366} & \textbf{22.34} & \textbf{148.2} \\
        \bottomrule
    \end{tabular}
    \label{tab:ablation_davis}
\end{table}

\subsection{Qualitative Result}
We present the qualitative results, as illustrated in Fig. \ref{fig:quality}. Notably, when the first and last frames involve non-linear transformations or large motion, traditional methods such as AMT and FILM fail to generate accurate intermediate frames. In contrast, our model generates superior results, characterized by accurate motion and clear textures. Furthermore, our method outperforms SVDKFI in low-resolution scenarios, demonstrating greater robustness to the effects of down sampling. Furthermore, our approach exhibits the capability to interpolate across diverse image styles. We evaluate its performance in real-world, anime, and sketch styles, as illustrated in Fig. \ref{fig:style}.



\subsection{Ablation study}
We conducted experiments on the proposed components of our method to perform an ablation study. Detailed results are provided in Table \ref{tab:ablation_davis}. To mitigate the impact of down sampling on model performance, we discard the \(256 \times 256\) version of DAVIS-7. Instead, we re-crop the DAVIS 2017 dataset to a resolution of \(512 \times 512\), retaining sequences of 9 frames, where the model is tasked with predicting the middle 7 frames. The baseline model includes only the spatial branch, enabling an image-to-video model to adapt to video frame interpolation. In contrast, our method integrates the temporal branch and the cross-frame attention module. Experimental results demonstrate that the proposed components significantly enhance the performance of the baseline model.

\section{Limitation and Discussion}
There are several limitations associated with the Stable Video Diffusion model. Firstly, Stable Video Diffusion often struggles to generate motion-rich videos, which means it performs poorly in cases requiring interpolation of ambiguous motion. Secondly, as a latent diffusion model, it down-samples the input into a latent space, which can compromise fine-grained details and adversely affect pixel-level metrics such as PSNR and SSIM. Additionally, Stable Video Diffusion has its own inherent limitations in output resolution, restricting its applicability in real-world scenarios.

\section{Conclusion}
We introduce a conditional encoder designed to adapt an image-to-video model for large-motion frame interpolation. To enhance performance, we integrate a dual-branch feature extractor and propose a cross-frame attention mechanism that effectively captures both spatial and temporal information, enabling accurate interpolation of intermediate frames. Despite its simplicity, our method proves highly effective, achieving comparable results on the Fréchet Video Distance (FVD) metric when evaluated against other state-of-the-art approaches.

\bibliographystyle{IEEEtran}
\bibliography{reference}

\begin{thebibliography}{10}
\providecommand{\url}[1]{#1}
\csname url@samestyle\endcsname
\providecommand{\newblock}{\relax}
\providecommand{\bibinfo}[2]{#2}
\providecommand{\BIBentrySTDinterwordspacing}{\spaceskip=0pt\relax}
\providecommand{\BIBentryALTinterwordstretchfactor}{4}
\providecommand{\BIBentryALTinterwordspacing}{\spaceskip=\fontdimen2\font plus
\BIBentryALTinterwordstretchfactor\fontdimen3\font minus \fontdimen4\font\relax}
\providecommand{\BIBforeignlanguage}[2]{{%
\expandafter\ifx\csname l@#1\endcsname\relax
\typeout{** WARNING: IEEEtran.bst: No hyphenation pattern has been}%
\typeout{** loaded for the language `#1'. Using the pattern for}%
\typeout{** the default language instead.}%
\else
\language=\csname l@#1\endcsname
\fi
#2}}
\providecommand{\BIBdecl}{\relax}
\BIBdecl

\bibitem{Cheng2020VideoFI}
X.~Cheng and Z.~Chen, ``Video frame interpolation via deformable separable convolution,'' in \emph{AAAI Conference on Artificial Intelligence}, 2020.

\bibitem{Danier2021STMFNetAS}
D.~Danier, F.~Zhang, and D.~R. Bull, ``St-mfnet: A spatio-temporal multi-flow network for frame interpolation,'' \emph{Computer Vision and Pattern Recognition}, pp. 3511--3521, 2021.

\bibitem{Jiang2017SuperSH}
H.~Jiang, D.~Sun, V.~Jampani, M.-H. Yang, E.~G. Learned-Miller, and J.~Kautz, ``Super slomo: High quality estimation of multiple intermediate frames for video interpolation,'' \emph{Computer Vision and Pattern Recognition}, pp. 9000--9008, 2017.

\bibitem{Niklaus2020SoftmaxSF}
S.~Niklaus and F.~Liu, ``Softmax splatting for video frame interpolation,'' \emph{Computer Vision and Pattern Recognition}, pp. 5436--5445, 2020.

\bibitem{Sim2021XVFIEV}
H.~Sim, J.~Oh, and M.~Kim, ``Xvfi: extreme video frame interpolation,'' \emph{International Conference on Computer Vision}, pp. 14\,469--14\,478, 2021.

\bibitem{Kong2022IFRNetIF}
L.~Kong, B.~Jiang, D.~Luo, W.~Chu, X.~Huang, Y.~Tai, C.~Wang, and J.~Yang, ``Ifrnet: Intermediate feature refine network for efficient frame interpolation,'' \emph{Computer Vision and Pattern Recognition}, pp. 1959--1968, 2022.

\bibitem{LDMVFI}
D.~Danier, F.~Zhang, and D.~R. Bull, ``Ldmvfi: Video frame interpolation with latent diffusion models,'' in \emph{AAAI Conference on Artificial Intelligence}, 2023.

\bibitem{VIDM}
S.~Jain, D.~Watson, E.~Tabellion, A.~Holy'nski, B.~Poole, and J.~Kontkanen, ``Video interpolation with diffusion models,'' \emph{Computer Vision and Pattern Recognition}, pp. 7341--7351, 2024.

\bibitem{SVDKFI}
X.~Wang, B.~Zhou, B.~Curless, I.~Kemelmacher-Shlizerman, A.~Holynski, and S.~M. Seitz, ``Generative inbetweening: Adapting image-to-video models for keyframe interpolation,'' \emph{arXiv preprint arXiv:2408.15239}, 2024.

\bibitem{framer}
W.~Wang, Q.~Wang, K.~Zheng, H.~Ouyang, Z.~Chen, B.~Gong, H.~Chen, Y.~Shen, and C.~Shen, ``Framer: Interactive frame interpolation,'' \emph{arXiv preprint arXiv:2410.18978}, 2024.

\bibitem{controlnet}
L.~Zhang, A.~Rao, and M.~Agrawala, ``Adding conditional control to text-to-image diffusion models,'' in \emph{International Conference on Computer Vision)}, Oct. 2023.

\bibitem{Attention}
A.~Vaswani, N.~M. Shazeer, N.~Parmar, J.~Uszkoreit, L.~Jones, A.~N. Gomez, L.~Kaiser, and I.~Polosukhin, ``Attention is all you need,'' in \emph{Advances in Neural Information Processing Systems}, 2017.

\bibitem{AMT}
Z.~Li, Z.-L. Zhu, L.~Han, Q.~Hou, C.~Guo, and M.-M. Cheng, ``Amt: All-pairs multi-field transforms for efficient frame interpolation,'' \emph{Computer Vision and Pattern Recognition}, pp. 9801--9810, 2023.

\bibitem{FILM}
F.~Reda, J.~Kontkanen, E.~Tabellion, D.~Sun, C.~Pantofaru, and B.~Curless, ``Film: Frame interpolation for large motion,'' in \emph{European Conference on Computer Vision}.\hskip 1em plus 0.5em minus 0.4em\relax Springer, 2022, pp. 250--266.

\bibitem{RIFE}
Z.~Huang, T.~Zhang, W.~Heng, B.~Shi, and S.~Zhou, ``Rife: Real-time intermediate flow estimation for video frame interpolation,'' \emph{ArXiv}, vol. abs/2011.06294, 2020.

\bibitem{ddpm}
J.~Ho, A.~Jain, and P.~Abbeel, ``Denoising diffusion probabilistic models,'' in \emph{Advances in Neural Information Processing Systems}, Dec. 2020.

\bibitem{ddim}
J.~Song, C.~Meng, and S.~Ermon, ``Denoising diffusion implicit models,'' in \emph{International Conference on Learning Representations}, Jan. 2021.

\bibitem{diffusion_beat_gan}
P.~Dhariwal and A.~Nichol, ``Diffusion models beat gans on image synthesis,'' in \emph{Advances in Neural Information Processing Systems}, Dec. 2021.

\bibitem{cfg}
J.~Ho and T.~Salimans, ``Classifier-free diffusion guidance,'' in \emph{NeurIPS 2021 Workshop on Deep Generative Models and Downstream Applications}, Dec. 2021.

\bibitem{ldm}
R.~Rombach, A.~Blattmann, D.~Lorenz, P.~Esser, and B.~Ommer, ``High-resolution image synthesis with latent diffusion models,'' in \emph{Computer Vision and Pattern Recognition}, Jun. 2022.

\bibitem{SVD}
A.~Blattmann, T.~Dockhorn, S.~Kulal, D.~Mendelevitch, M.~Kilian, and D.~Lorenz, ``Stable video diffusion: Scaling latent video diffusion models to large datasets,'' \emph{ArXiv}, vol. abs/2311.15127, 2023.

\bibitem{CogVideo}
W.~Hong, M.~Ding, W.~Zheng, X.~Liu, and J.~Tang, ``Cogvideo: Large-scale pretraining for text-to-video generation via transformers,'' in \emph{International Conference on Learning Representations}, 2023.

\bibitem{animediff}
Y.~Guo, C.~Yang, A.~Rao, Z.~Liang, Y.~Wang, Y.~Qiao, M.~Agrawala, D.~Lin, and B.~Dai, ``Animatediff: Animate your personalized text-to-image diffusion models without specific tuning,'' \emph{International Conference on Learning Representations}, 2024.

\bibitem{Align}
A.~Blattmann, R.~Rombach, H.~Ling, T.~Dockhorn, S.~W. Kim, S.~Fidler, and K.~Kreis, ``Align your latents: High-resolution video synthesis with latent diffusion models,'' \emph{Computer Vision and Pattern Recognition}, pp. 22\,563--22\,575, 2023.

\bibitem{make-a-video}
U.~Singer, A.~Polyak, T.~Hayes, X.~Yin, J.~An, S.~Zhang, Q.~Hu, H.~Yang, O.~Ashual, O.~Gafni, D.~Parikh, S.~Gupta, and Y.~Taigman, ``Make-a-video: Text-to-video generation without text-video data,'' in \emph{International Conference on Learning Representations}, 2023.

\bibitem{sparse}
Y.~Guo, C.~Yang, A.~Rao, M.~Agrawala, D.~Lin, and B.~Dai, ``Sparsectrl: Adding sparse controls to text-to-video diffusion models,'' \emph{ArXiv}, 2023.

\bibitem{clip}
A.~Radford, J.~W. Kim, C.~Hallacy, A.~Ramesh, G.~Goh, S.~Agarwal, G.~Sastry, A.~Askell, P.~Mishkin, J.~Clark \emph{et~al.}, ``Learning transferable visual models from natural language supervision,'' in \emph{International Conference on Machine Learning}, Jul. 2021.

\bibitem{RAFT}
Z.~Teed and J.~Deng, ``Raft: Recurrent all-pairs field transforms for optical flow,'' in \emph{European Conference on Computer Vision}, 2020.

\bibitem{Depth}
R.~Ranftl, K.~Lasinger, D.~Hafner, K.~Schindler, and V.~Koltun, ``Towards robust monocular depth estimation: Mixing datasets for zero-shot cross-dataset transfer,'' \emph{IEEE Transactions on Pattern Analysis and Machine Intelligence}, vol.~44, pp. 1623--1637, 2019.

\bibitem{CBAM}
S.~Woo, J.~Park, J.-Y. Lee, and I.~S. Kweon, ``Cbam: Convolutional block attention module,'' in \emph{Proceedings of the European conference on computer vision (ECCV)}, 2018, pp. 3--19.

\bibitem{webvid10m}
M.~Bain, A.~Nagrani, G.~Varol, and A.~Zisserman, ``Frozen in time: A joint video and image encoder for end-to-end retrieval,'' \emph{International Conference on Computer Vision}, pp. 1708--1718, 2021.

\bibitem{vimeo90k}
T.~Xue, B.~Chen, J.~Wu, D.~Wei, and W.~T. Freeman, ``Video enhancement with task-oriented flow,'' \emph{International Journal of Computer Vision}, vol. 127, no.~8, pp. 1106--1125, 2019.

\bibitem{DAVIS}
J.~Pont-Tuset, F.~Perazzi, S.~Caelles, P.~Arbel{\'a}ez, A.~Sorkine-Hornung, and L.~V. Gool, ``The 2017 davis challenge on video object segmentation,'' \emph{ArXiv}, 2017.

\bibitem{UCF101}
K.~Soomro, A.~Zamir, and M.~Shah, ``Ucf101: A dataset of 101 human actions classes from videos in the wild,'' \emph{ArXiv}, 2012.

\bibitem{lpips}
R.~Zhang, P.~Isola, A.~A. Efros, E.~Shechtman, and O.~Wang, ``The unreasonable effectiveness of deep features as a perceptual metric,'' in \emph{Computer Vision and Pattern Recognition}, Jun. 2018.

\bibitem{fid}
M.~Heusel, H.~Ramsauer, T.~Unterthiner, B.~Nessler, and S.~Hochreiter, ``Gans trained by a two time-scale update rule converge to a local nash equilibrium,'' in \emph{Advances in Neural Information Processing Systems}, Dec. 2017.

\bibitem{fvd}
S.~Ge, A.~Mahapatra, G.~Parmar, J.-Y. Zhu, and J.-B. Huang, ``On the content bias in fréchet video distance,'' in \emph{Computer Vision and Pattern Recognition}, 2024.

\end{thebibliography}
\end{document}